# A Multiresolution Clinical Decision Support System Based on Fractal Model Design for Classification of Histological Brain Tumours

Omar S. Al-Kadi*

Institute of Biomedical Engineering

University of Oxford, OX3 7DQ United Kingdom

*Abstract*— Tissue texture is known to exhibit a heterogeneous or non-stationary nature; therefore using a single resolution approach for optimum classification might not suffice. A clinical decision support system that exploits the subbands' textural fractal characteristics for best bases selection of meningioma brain histopathological image classification is proposed. Each subband is analysed using its fractal dimension instead of energy, which has the advantage of being less sensitive to image intensity and abrupt changes in tissue texture. The most significant subband that best identifies texture discontinuities will be chosen for further decomposition, and its fractal characteristics would represent the optimal feature vector for classification. The performance was tested using the support vector machine (SVM), Bayesian and *k*-nearest neighbour (kNN) classifiers and a leave-one-patient-out method was employed for validation. Our method outperformed the classical energy based selection approaches, achieving for SVM, Bayesian and kNN classifiers an overall classification accuracy of 94.12%, 92.50% and 79.70%, as compared to 86.31%, 83.19% and 51.63% for the co-occurrence matrix, and 76.01%, 73.50% and 50.69% for the energy texture signatures; respectively. These results indicate the potential usefulness as a decision support system that could complement radiologists' diagnostic capability to discriminate higher order statistical textural information; for which it would be otherwise difficult via ordinary human vision.

## I. INTRODUCTION

Meningioma tumours usually occur in adults, with a marked female bias represented by a one to three man to women ratio [1]. It also accounts for 27% of all primary brain tumours, making it the most common tumour of that type [2]. An automated meningioma diagnosis system is essential in improving reproducibility by overcoming subjective diagnosis due to variability associated with expert's evaluation. That is, when differences become minor between tumour subtypes of the same grade this might trigger for an increase in intra-observer variability, i.e. pathologist not being able to give the same reading of the same image at more than one occasion, and inter-observer variability, i.e. increase in classification variation between different pathologists; and thereby increasing uncertainty that may impact patient outcome. While the dataset could be acquired at fixed magnification and microscope settings, and with the same staining protocol, the variability in the reported diagnosis may still occur [3, 4]. This could be attributed to the non-homogeneous nature of the diseases, namely, not all samples referring to a certain tumour subtype look identical; raising the issue of misclassification. The automated diagnosis system could also assist in overcoming other diagnosis variability-related subjective factors such as, preconception, expectations, relying on diligence, and fatigue, which could cause differences in image perception. Meningiomas can have three grades numbered from I till III, over here we are more concerned with classifying different subtypes within the same grade, which is considered a more challenging task compared to grade classification, as histopathological features tend to become easier to differentiate by the naked eye in the latter compared to the former case.

*E-mail o.alkadi@ju.edu.jo; omar.al-kadi@eng.ox.ac.uk.
 Tel: (+44) 1865 617722.

The main concern in the texture analysis problem is two-fold: to capture distinctive characteristics that will maximise the difference between the various image regions, and the selection of the best pattern classifier that can give the optimal performance. Some examples of the different methods used for cytopathological diagnosis include immunocytochemistry, cytophotometry, flow cytometry, and microarray-based comparative genomic hybridisation. However, nuclear texture or chromatin granularity is also considered an important measure in cytopathological diagnosis for its capability of characterising disease or abnormality in cell structures, based on the assumption that different tumour grades tend to form different shapes and behaviours and hence reflect on the tissue texture general appearance; and thus could be easily characterised and captured by a machine learning algorithm. Moreover, approaching the problem for a texture analysis perspective can provide the capability of discriminating higher order statistical textural information for which it would be otherwise difficult via ordinary human vision. However, histopathological tissue texture is known to be heterogeneous, and a varying degree of texture heterogeneity exists. The non-stationary nature of this kind of texture weakens the ability for an effective automated classification from a monoresolution viewpoint, and image pre-processing prior to feature extraction might not suffice. On the other hand, viewing texture from a multiresolution perspective can filter out irrelevant features and noise while simultaneously giving more emphasis on the features that contribute to better distinction. Techniques such as wavelet transform can break down textures' statistical complexity to distinguish between different texture regions, and their high sensitivity to local features facilitates the processes of preattentive or subtle texture discrimination as well [5]. Furthermore, according to the uncertainty principle, the wavelet transform can achieve an optimal joint spatial-frequency localisation, i.e. simultaneously maintain a good boundary accuracy and frequency response [6].

Wavelet packet (WP) is a generalised framework of the multiresolution analysis and comprises all possible combination of subbands decomposition. However, it is unwieldy to use all frequency subbands for texture characterisation as not all of them have the same discriminating power, and inclusion of weak subbands would have a negative impact on the classifier's performance. Whereas using an exhaustive search would be computational expensive as the number of decomposition levels grows higher. Therefore an adaptive approach is required for selection of the basis with prominent discriminating power.

## II. Overview of Previous Work

The WP subbands selection can be performed either by selecting the best bases from a library of WPs or in a tree-structured approach. Coifman and Wickerhauser proposed to choose the best basis which gave the most compact representation after transforming the signal into different WP bases [7]. The entropy was used as the cost function for selection of the decomposition levels, where the subband that minimises the cost function, from a comparison between the nodes and its leaves in the WP decomposition tree, was considered the optimal choice. By extending the additive cost function in [7] to an arithmetic hence a geometric mean, Dansereau et al proposed a generalised rényi entropy for best basis search [8], allowing for different moment orders and inclusion of possible incomplete probabilities in the search as well. Saito et al estimated the probability density of each class in each coordinate in the WP and local trigonometric bases, then applied the Kullback–Leibler divergence as a distance measure among the densities for selection of the most discriminating coordinates [9]. While Rajpoot compared the discrimination energy between the subbands by using four different distance metrics [10]. The Kullback-Leibler divergence, Jensen-Shannon divergence, Euclidean distance, and Hellinger distance were used to assess the dissimilarities in-between the WPs for selection of the most discriminant bases. Others excluded the set of frequency subbands whose energy signatures showed a degree of dependence identified by mutual information [11]. Huang and Aviyente developed an algorithm for subband selection based on the dependency of the extracted energy features. A compact feature representation was achieved when the dependence between the subbands and the evaluated score of individual subbands was incorporated [12]. Another related work was based on best clustering bases, wherein clustering basis functions are selected according to their ability to separate the fMRI time series into activated and non-activated clusters [13]. The basis that concentrates the most discriminatory power on a small number of basis functions is selected. On the other hand, a tree-structured technique for best basis selection was proposed by Chang and Kuo, where only the subbands with the highest energy are selected for further decomposition [14]. An averaged $l_1$ – norm was used as the energy function for location of the dominant frequency channels, and decomposition is stopped if subbands' energy is less than a factor of the maximum energy at that resolution. Acharyya and Kundu used $M$-band WP decomposition that resulted in a large number of subbands, therefore they decomposed the subbands whose total energy value are greater than the energy of all subbands at the same resolution [15].

Regarding application of WPs to meningiomas, Lessmann et al employed a self-organising map in order to link the morphological histopathological image characteristics to the space spanned by features derived from hue, saturation and brightness colour model and WP transform [16]. For four different subtypes of meningiomas, an average of 79% for the entire dataset was classified correctly. Also Wirjadi et al applied a supervised learning method for classification of meningioma cells [17], using a decision tree, the most relevant features were selected from a base of grey and coloured image features. Others extracted features from four meningioma subtypes using adaptive WP transform and local binary patterns (LBP) methods [18] [19]. In the applied WP technique, the best sets of subbands are selected by simply measuring the discriminating power between all decomposed subbands at a certain level using Hellinger distance. While for the LBP method, first order statistics were derived from its histogram; both studies reported a classification accuracy of 82.1%.

Having a relatively good feature extraction method that can characterise the underlying physiology of the examined tissue texture would be considered a half way through designing a robust histopathological meningioma subtype decision support system. The other necessary step that could complete the picture is choosing an appropriate machine-learning algorithm that can assist in improving the classification accuracy, when the quality of the extracted features might degrade due to tissue heterogeneity. In our earlier work [20, 21], we proposed a different approach for best basis selection for the processes of histopathological meningioma classification, and in this work we develop a clinical decision support system based on an improvement of the best basis selection method and further integrate it with the optimal classifier model design which would yield more significantly sensitive and specific results. Major additional technical details in this paper are as follows:

   i) The fractal model design used in the wavelet packet decomposition was described in details and an algorithm for implementation was provided. The algorithm was modified as well to include the computation of the texture lacunarity to work along with the fractal dimension for determination of the optimal decomposition levels. Also an example was provided that illustrates the capability of higher frequency channels of meningioma to provide stronger Fourier spectrum, and hence more information.

   ii) The selection of the lambda threshold, which determines how deep the image resolution can be probed, was experimentally justified and results illustrated.

   iii) The resulting subbands spanned from our best basis selection method were indicated for all four meningioma subtypes, and further illustrated via classification accuracies for each decomposition level.

iv) Since a feature set may be performing better than others only because its distribution is a better fit for the assumptions underlying the classifier model, the proposed method classification performance was designed and implemented using two other classifiers, namely SVM and kNN in addition to the previously reported NBC, in order to investigate the classifiers' influence on the classification accuracy for a given feature set provided by our method. Also, SVM ensembles with bagging were applied for improving reported accuracies and a leave-one-patient-out validation technique was employed instead of a hold-out for evaluation of performance.

v) Accuracies using all three classifiers were compared, and confusion matrices and sensitivity and specificity tests were presented. Also the number of support vectors for the SVM classifier that performed best was presented for demonstration of possible generalisation capacity.

vi) The proposed technique was further compared with non-tissue synthetic Brodatz textures and further compared with other established texture analysis techniques for broadening the usefulness and proving of robustness.

In this work the fractal dimension (FD) is used for guiding the subband tree-structure decomposition instead of energy that is highly dependent on the subband intensity, and the subbands absolute difference combined with another parameter in the fractal model that measures the heterogeneity of the tissue were used to automatically terminate the WP decomposition at the optimal resolution levels. The feature vector will consist of FDs of all optimal selected subbands according to their fractal discriminating power, acting as a feature reduction technique. The motivation to use such texture measure, besides its scale invariance or the capability to investigate self-similarity, is its surface roughness estimation that can be used to detect variance in cell nucleus structure orientation and size for differentiating between meningioma subtypes. Fractal analysis for the purpose of tumour discrimination at a micro-scale was proven to be successful in numerous studies related to various medical imaging modalities as in computed tomography [22], X-ray [23], magnetic resonance [24], and ultrasound [25]. This work takes advantage of FD in diagnosing histopathological texture, and applies it at a macro-scale for images acquired by digital microscopy modality. Also with the large size of the meningioma images, the tree-structure was favoured to reduce computational time in order to explore the full texture characteristics at deeper levels. Moreover, overcomplete dyadic wavelet transform was used, holding the size of the transformed image to be the same as the original image without any down-sampling. The proposed method was tested using three different classifiers to investigate its robustness and followed by statistical analysis to measure its significance. Fig.1 illustrates the technique developed in this work.

This paper lays the foundations of wavelet theory in section II, followed by image acquisition and pre-processing, followed by the developed approach for best subband selection in section III and IV, respectively. The texture feature classification approach is discussed in section V, and the results and its interpretation in section VI and VII, respectively. Finally, the paper concludes in section VIII.

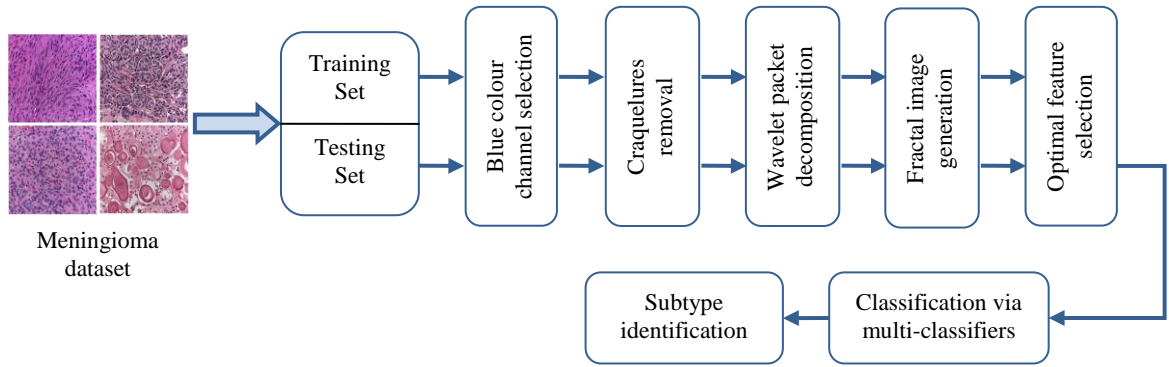

Fig. 1. Diagram explaining the main processes for the developed brain tumour decision support system.

## III. TIME-FREQUENCY LOCALISATION

Multiresolution or multi-scale analysis is a fine to coarse analysis strategy for which the signal details are decomposed and examined at different levels of resolution. In terms of pattern recognition, large structures or high contrast are best localised at low resolution levels, while higher levels would be more appropriate for small size or low contrast objects. Therefore multiresolution processing gives the advantage of analysing both small and large object characteristics in a single image at several resolutions. The decomposition of the image into multiple resolutions based on small basis functions of varying frequency and limited duration called wavelets was first introduced by Mallat [26], which is also discussed in detail in [27-29]. The wavelet analysis approach can be regarded as the scale $j$ and translation $k$ of a basic function (called a mother wavelet) to cover the time-frequency domain. The discrete wavelet transform for a function $f(x,y)$ of size $M \times N$ can be represented as follows [30]:

$$W_\varphi(j_0,m,n) = \frac{1}{\sqrt{MN}} \sum_{x=0}^{M-1} \sum_{y=0}^{N-1} f(x,y)\, \varphi_{j_0,m,n}(x,y) \qquad (1)$$

$$W_\psi^i(j,m,n) = \frac{1}{\sqrt{MN}} \sum_{x=0}^{M-1} \sum_{y=0}^{N-1} f(x,y)\, \psi_{j,m,n}^i(x,y) \qquad (2)$$

where $f(x, y) \in L^2(\Re)$ relative to scaling $\varphi(x,\ y)$ and wavelet function $\psi(x,\ y)$, and $W_\varphi(j_0,m,n)$ defines an approximation of $f(x,y)$ at scale $j_0$, and $W_\psi^i(j,m,n)$ coefficients add horizontal, vertical and diagonal details for scales $j \geq j_0$. By decomposing the signal's approximation coefficients as well, the wavelet transform can be extended in the middle and high frequency channels (LH, HL and HH-bands) and not only in the low frequency channels (LL-band), providing a better partitioning of the spatial-frequency domain, which is known as the WP transform [7]. As features of some textures would be more prevalent in the higher frequency channels, WPs would give the high frequency structures in an image an equal opportunity for investigation of possible interesting information. We are concerned more with a better representation of the texture characteristics at each decomposition; therefore, this work presents an *overcomplete* tree-structured wavelet packet representation by omitting image down-sampling operation at each decomposition step, holding the size of the transformed image the same as the original image.

Achieving a good spatial-frequency localisation depends on understanding the relation between the two domains. It is known that the time and frequency are inversely related according to the Heisenberg uncertainty principle. This means that more time domain precision in analysing a certain function will be at the expense of frequency precision, and vice versa. Thus having a varying size and constant area window which adapts to the range of the analysed frequencies could manage the trade-off. This is illustrated graphically in Fig. 2 using tiles, also called Heisenberg boxes, which show the concentration of the basis functions' energy [28, 31, 32]. Orthonormal basis functions are assumed, therefore time-frequency tiling is characterised by non-overlapping tiles. Time-frequency tiling using a delta function basis as in Fig. 2(a) identifies the time of sampling occurrence but provides no frequency information. Inversely, the Fourier transform in Fig. 2(b) provides the frequency information but lacks the time resolution. The windowed Fourier transform decomposes the signal into a set of equal size frequency intervals resulting in constant frequency and time resolution, as shown in Fig. 2(c). Finally, the wavelet transform has varying size frequency intervals that can pack all oscillations of the basis wavelet into a narrow interval for high frequencies and into wide intervals for low frequencies, shown in Fig. 2(d). This way non-periodic and/or non-stationary functions whose frequencies vary in time can be more appropriately analysed. As short or narrow basis functions are required for detection of signal discontinuities, detailed frequency analysis requires long or wide basis functions. The wavelet transform using a Daubechies wavelet basis functions for example can represent both situations by having narrow high and wide low frequencies simultaneously, see Fig. 3(b). Since this work is mainly concerned with characterising the fractal characteristics of tissue texture, and the Daubechies wavelet family is known for its capability in detecting and representing signals that exhibit fractal patterns [28]; therefore it was the most suitable choice from the other known wavelet families for this work. Besides, it is also considered to be sensitive in recognition of fine characteristic structures, and its application of overlapping windows, unlike other wavelets such as the Haar wavelet, facilitates the capture of all high frequency changes easily.

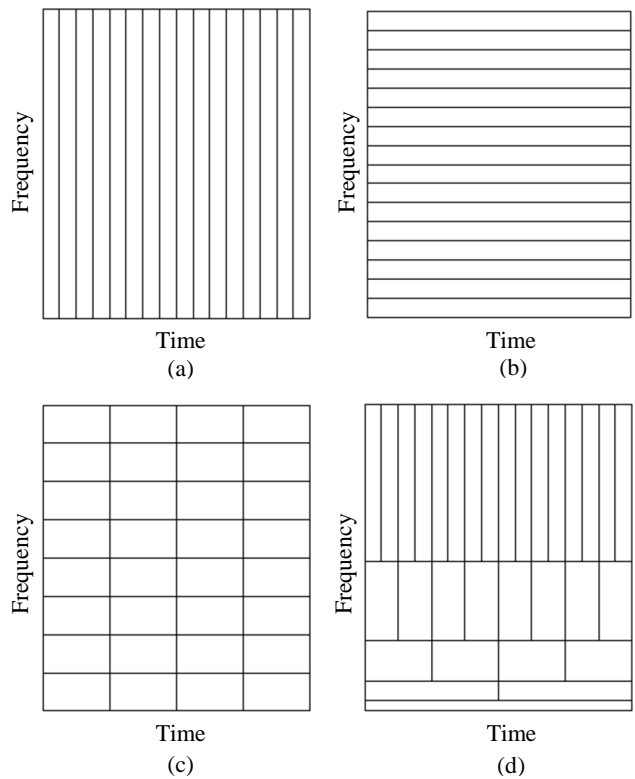

Fig. 2 Time-frequency tiling for (a) sampled data, (b) Fourier transform, (c) windowed Fourier transform, and (d) wavelet transform.

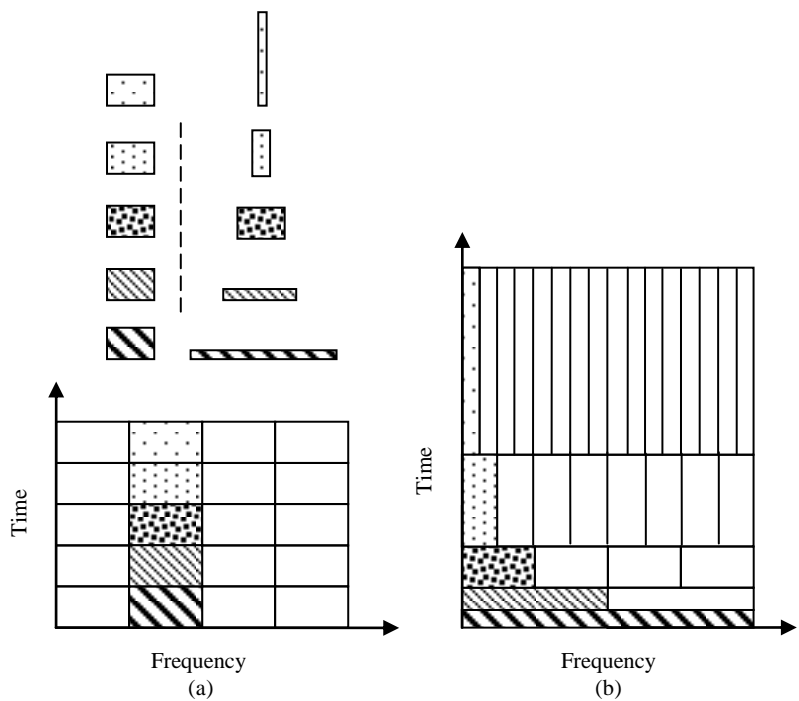

Fig. 3 Comparing tiling covering of the time-frequency domain using (a) sinusoidal basis function for windowed Fourier transform, and (b) Daubechies wavelet basis functions for wavelet transform.

## IV.  IMAGE PREPARATION

### A.  Acquisition stage

Four subtypes of grade I meningioma tissue biopsies distinguished according to the World Health Organisation grading system [33] are used in this study, shown in Fig. 4. Each subtype has its own textural features as shown in Table I, which pathologists look for in the processes of tumour classification [34, 35]. The diagnostic tumour samples were derived from neurosurgical resections at the Bethel Department of Neurosurgery, Bielefeld, Germany for therapeutic purposes, routinely processed for formalin fixation and embedded into paraffin.

Four micrometer thick microtome sections were dewaxed on glass slides, stained with Mayer's haemalaun and eosin (H&E), dehydrated and cover-slipped with mounting medium (Eukitt®, O. Kindler GmbH, Freiburg, Germany). Archive material of cases from the years 2004 and 2005 were selected to represent typical features of each meningioma subtype. Slides were analysed on a Zeiss Axioskop 2 plus microscope with a Zeiss Achroplan 40×/0.65 oil immersion objective. After manual focusing and automated background correction, images were taken at a resolution of 1300 × 1030 pixels, 24 bits, true colour RGB at standardised 3200 K light temperature in TIF format using Zeiss AxioVision 3.1 software and a Zeiss AxioCam HRc digital colour camera (Carl Zeiss AG, Oberkochen, Germany). Five typical cases were selected for each diagnostic group and four different photomicrographs were taken of each case, resulting in a set of 80 pictures. Each original picture was truncated to 1024 × 1024 pixels and then subdivided in a 2 × 2 subset of 512 × 512 pixel pictures. This resulted in a database of 320 sub-images for further analysis. All acquired images were fully anonymised and this work did not influence the diagnostic process or the patients' treatment.

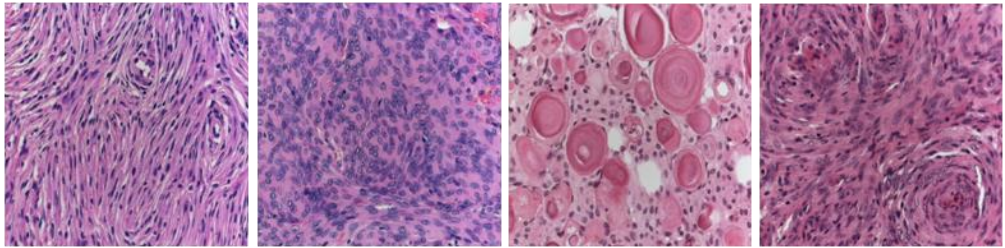

Fig. 4. (left-right) Meningioma fibroblastic, meningothelial, psammomatous and transitional subtypes.

TABLE I MAIN HISTOLOGICAL TEXTURAL FEATURES FOR THE FOUR MENINGIOMA SUBTYPES

| Subtype | Characteristics |
|---------|-----------------|
| Fibroblastic | Spindle-shaped cells resembling fibroblasts in appearance, with abundant amounts of pericellular collagen. |
| Meningothelial | Broad sheets or lobules of fairly uniform cells with round or oval nuclei. |
| Psammomatous | A variant of transitional meningiomas with abundant psammoma bodies and many cystic spaces. |
| Transitional | Contains whorls, few psammoma bodies and cells having some fibroblastic features (i.e. spindle-shaped cells) |

The relation between the fractal features and the underlying biological features manifest the discrimination between the four subtypes. Distinctive characteristics such as the small whirlpools in the form of spinning body of nuclei related to the transitional and appearing subtle in the fibroblastic subtypes; the round collection of calcium, known as psammoma bodies, that occurs repeatedly and in varying sizes in the psammomatous subtype; and the enlarged oval shape nuclei in the meningothelial which are condensed in approximate clusters, all present potential fractal characteristics that can be exploited for optimal feature extraction to facilitate the next stage of the classification process.

### B. Pre-processing stage

All four meningioma images were first configured prior to feature extraction by selecting the colour channel that would assist in better defining the boundary of the bluish colour cell nuclei from the pink cytoplasm background, therefore the blue colour channel from the RGB colour space was selected. This was followed by applying the morphological gradient $M_g$ to extract the general structure and reduction of possible mechanical distortion in the form of *craquelures* (appearing as white cracks) that may occur during biopsy preparation procedure. The $M_g$, as defined in (3), represents the difference between the dilation and erosion of an image, assisting in highlighting the size and orientation of the cell nuclei structure, which would reflect on the quality of the texture signatures to be extracted from each subband.

$$M_g = (I(x, y) \oplus k(x, y)) - (I(x, y) \ominus k(x, y)) \quad (3)$$

Here $I(x, y)$ is the image operated on by the structuring element $k(x, y)$ chosen empirically to be a square of 5×5 pixels of ones. Hence to improve the probabilistic estimate of the different subtypes, it is necessary to denoise mechanical distortion that may affect the specific characteristics of the cell nuclei distribution before the feature extraction phase.

## V. SUBBAND SELECTION OPTIMISATION

A 8-tap Daubechies filter [28, 36] (see Table II) – which is widely used in characterising signals exhibiting fractal patterns [28] – was used to obtain the WPs and decomposition was implemented in a tree structure approach [14], expanding only the basis having the most significant signature. This approach was adopted to investigate the possibility of higher frequency channels to provide significant information as compared to the classical low-frequency decomposition approach for the processed histopathological images. Fig. 5 supports this trend, in which the middle and high wavelets subbands for the first level of decomposition have stronger Fourier spectrum as compared to the low frequency channel, especially for the first level of resolution.

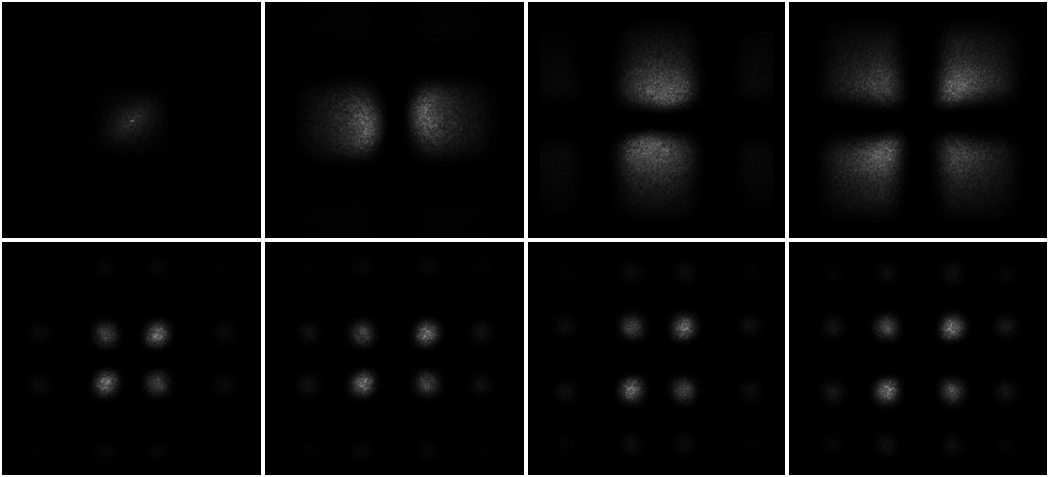

Fig. 5. Fibroblastic meningioma power spectrum of LL, LH, HL and HH wavelet bands for the 1$^{st}$ (upper row) and 3$^{rd}$ (lower row) level of resolution.

TABLE II
WAVELET TRANSFORM 8-TAP
DAUBECHIES FILTER COEFFICIENTS

| Coefficient | Value |
|---|---|
| $h(0)$ | 0.16290184 |
| $h(1)$ | 0.50547316 |
| $h(2)$ | 0.44610023 |
| $h(3)$ | -0.01978767 |
| $h(4)$ | -0.13225371 |
| $h(5)$ | 0.02180788 |
| $h(6)$ | 0.02325179 |
| $h(7)$ | -0.00749321 |

*A. Fractal model design*

The wavelet subband selection process was based upon exploiting possible fractal characteristics that images of meningioma cell nuclei may possess. There are several fractal models used to estimate the fractal dimension (FD). The fractal Brownian motion that is the mean absolute difference of pixel pairs as a function of scale and also known to perform well in case of sparse data was adopted [37], which can be defined as

$$E(\Delta I) = v \Delta r^H \qquad (4)$$

Herein, $\Delta I = \mid I(x_2, y_2) - I(x_1, y_1) \mid$ is the mean absolute difference of pixel pairs; $\Delta r = [(x_2 - x_1)^2 + (y_2 - y_1)^2]^{1/2}$ is the pixel pair distances; $H$ is called the Hurst coefficient; and $v$ is a constant, $v$ with $> 0$. The FD can be then estimated by plotting both sides of (4) on a log–log scale (see Fig. 6), and $H$ will represent the slope of the curve that is used to estimate the FD as: FD = $3 - H$.

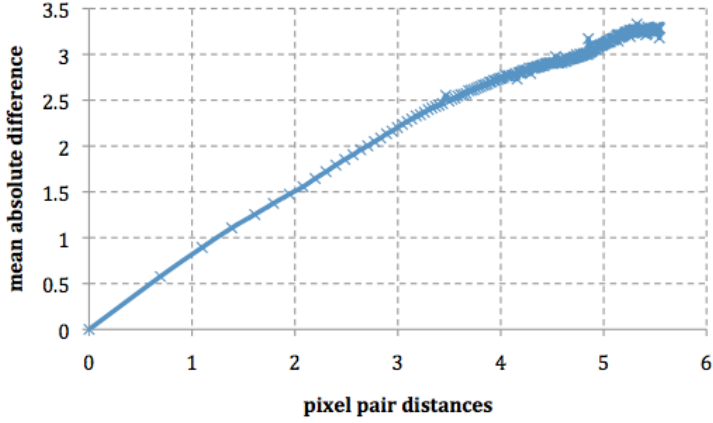

Fig. 6. Mean absolute difference ($\Delta I$) versus pixel pair distances ($\Delta r$) in log-log coordinates.

Each histopathological image $I(x, y)$ of size $M$ x $N$ is transformed to its corresponding FD version by applying the kernel $fd(s, t)$ of size $p$ x $q$ (a value of 7 was selected for both $p$ and $q$ [22]) as in (5). By operating by block processing on the neighboring pixels and finding the FD as described in the preceding paragraph, each pixel in the generated fractal image $\Im$ would represent the FD derived from the surrounding neighbours of the corresponding pixel in the original image; giving an estimated value in the range between 2 and 3 for each pixel. The kernel is calculated as in (5) and applied as in (6), where the two variables $a$ and $b$ are nonnegative integers which are computed in order to centre the odd-size kernel on each operating pixel in the original image.

$$fd(s,t) = 3 - \left( \frac{\log(\Delta I / v)}{\log(\Delta r)} \right) \qquad (5)$$

$$\Im(x, y) = \sum_{s=-a}^{a} \sum_{t=-b}^{b} fd(s,t) I(x+s, y+t) \qquad (6)$$

where $a = \text{ceil}\left(\dfrac{p-1}{2}\right)$ and $b = \text{ceil}\left(\dfrac{q-1}{2}\right)$

The fractal characteristics are estimated for all subbands at each level of WP decomposition, where the FD is computed on a pixel-by-pixel basis to producing a fractal image for each subband − where each pixel has its own FD value, and the rougher the surface is, the higher the FD values, and vice versa. Features at image boundaries are computed after assuming the image is mirror-like continually extended in both directions. Then the expressive FD would represent the average value of the generated fractal image $\Im$ as in (7), where $j$ is the subbands at a decomposition level $i$ for a certain subtype $k$, which would give a more reliable estimation as compared if a single FD was directly estimated from the whole subband. Finally the subband with the highest FD is selected for further decomposition.

$$W_{i,j}^{k} = \frac{1}{MN}\sum_{r=1}^{MN}\Im_{r} \qquad (7)$$

At the end of the feature extraction stage, a feature vector $f = \left(W_{1,1}^{k}, W_{1,2}^{k}, \dots W_{i,j}^{k}\right)$ consisting of all selected subbands fractal features $j$ to a certain decomposition level $i$ will be produced for each of the meningioma images $k$. In order to save processing time and when the differences among the fractal features become less significant, a designated threshold $\lambda \, \square$ would reduce the dimensionality of the extracted feature vector. By that, unnecessary decompositions are avoided, which could contribute towards a positive effect on the classifier's performance, and would further assist in determining the optimal features; a crucial part in facilitating the classifier's task. Given a certain meningioma subtype $k$ for a specific subband $j$ at a certain decomposition level $i$, where $i \in \{LL, LH, HL, HH\}$ bands, which can be expressed as if the condition $\left(\forall f_{j}^{i} \in \ max\,(W_{i,j}^{k})_{FD}\right) \leq \lambda$ is satisfied, then the decomposition should terminate. Therefore, the FD signatures' absolute difference $\mathcal{D}_{m} = \left|W_{i,j}^{k} - W_{i,j+1}^{k}\right|$ between all four wavelets subbands ($W_{i,LL}^{k}$, $W_{i,LH}^{k}$, $W_{i,HL}^{k}$ and $W_{i,HH}^{k}$) for a certain resolution level $i$ needs to be less than or equal to $\lambda$ ($\lambda = 0.012$ after being optimised over a range of reasonable values) before decomposition terminates.

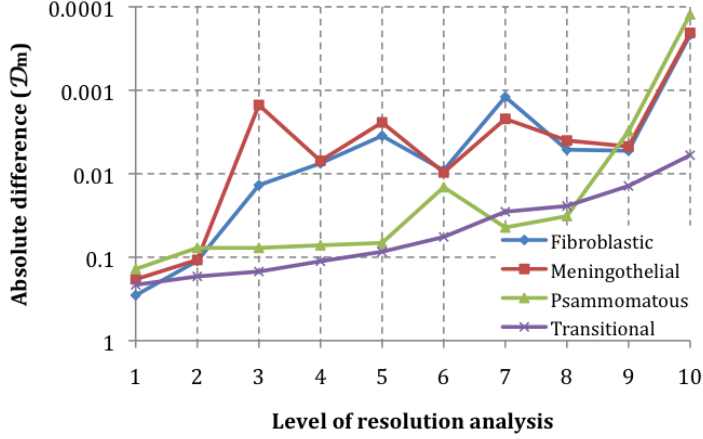

Fig. 7. Logarithmic plot for fractal dimension signature's absolute difference (D_m) variation down to 10 WP decomposition levels for all four meningioma subtypes. The set λ threshold was chosen to be at the value of 0.012 where all meningioma subtypes reach their optimal classification performance.

Fig. 7 shows that the meningioma subtypes reach their optimal classification performance - the results section will affirm this choice - before exceeding the 0.012 level; namely the third decomposition level for fibroblastic, second for meningothelial, eighth for psammomatous, and ninth for transitional subtypes.

Another parameter used in the designed fractal model that tends to enhance texture description and also provide a measurement of heterogeneity is the *lacunarity* ($\mathcal{L}_m$) defined as,

$$\mathcal{L}_m = \frac{1}{MN} \sum_{x=0}^{M-1} \sum_{y=0}^{N-1} \left| \frac{FD(x, y)}{1/MN \sum_{m=0}^{M-1} \sum_{n=0}^{N-1} FD(m, n)} - 1 \right| \qquad (8)$$

Usually, the $\mathcal{L}_m$ measure is used to further differentiate between two distinctive textures that might exhibit similar FD values. In our model, the lacunarity of the generated fractal image $\Im$ is computed instead of the original histopathological image $I$, and then used as an indication of heterogeneity, i.e. the lower the $\mathcal{L}_m$ value the more homogeneous the texture, and vice versa. Consequently, both the $\mathcal{D}_m$ and $\mathcal{L}_m$ would be used to terminate the WP decomposition when differences among the subbands become less significant, and thereby before the classifiers' accuracy begins to degrade again.

The best basis selection processes based on fractal analysis (which from now on will abbreviated to BBS$_{FD}$, for short) can be summarised in the following pseudo code:

---

**Algorithm** Best bases selection based on fractal wavelet packet decomposition.

**for** $k = 1$ to $l$ **do** % each meningioma subtype

  **for** $i = 1 \dots m$ **do** % each wavelet decomposition level

   **for** $j = 1 \dots n$ **do** % each filtered subband $W_{i,j}^{k}$

   Using a sliding box of size $p$ x $q$, generate a $M$ x $N$ fractal image for each subband $W_{i,j}^{k}$

    **for** $x = 1$ to $p$ **do**

     **for** $y = 1$ to $q$ **do**

    Find Hurst coefficient $H$;

    Calculate fractal dimension $FD$ $(x, y)$;

     **end for**

    **end for**

   Find the average of the generated fractal image for each decomposed subband

   **for** $x = 1$ to $M$ **do**

    **for** $y = 1$ to $N$ **do**

   $W_{i,j}^{k}$ = average $(FD)$ ;

   $\mathcal{L}_m$= lacunarity $(FD)$ ;

     **end for**

    **end for**

  **end for**

   Find the absolute difference and lacunarity between all FD signatures and compare with threshold

   $\forall$ ( $W_{i,j}^{k}$)$_{\text{FD}}$ subbands fractal signatures

   **if** $\left| W_{i,j}^{k} - W_{i,j+1}^{k} \right|$ OR $\mathcal{L}_m \leq \lambda$ **then**

    terminate and return optimal fractal features vector $f$;

   **else** $max( W_{i,j}^{k})$ choose the band with the highest FD signature for next level $l$ of decomposition;

   **end if**

  **end for**

**end for**

---

### B. Energy-based signatures

For the purpose of comparison, the performance of the FD would be benchmarked against other commonly used texture signatures for subband discriminant power assessment. These methods use the highest energy among the subbands for WPs expansion. Some of the most used methods for extraction of subbands texture signatures is the energy $E_k$ $k$, $k = 1$, $2$ in the form of $l_1$-norm [14] and $l_2$-norm [38] as in (9).

$$E_k = \frac{1}{MN} \sum_{x=0}^{M-1} \sum_{y=0}^{N-1} |I(x,y)|^k \qquad (9)$$

Here $M$ and $N$ are the size of the subband intensity $I(x, y)$, and we abbreviated this method as BBS$_E$. Also the co-occurrence matrix [39] $C_{\theta,\delta}(i, j)$ representing the joint probability of grey-level pixel $i$ and $j$, within a certain displacement distance $\delta$ and orientation θ. Then the unnormalised entry $C_{\theta,\delta}$ can be expressed as

$$C_{\theta,\delta}(i, j) = \#\left\{((x_1, y_1), (x_2, y_2)) \in (M \times N) \times (M \times N) :\right.$$
$$I(x_2, y_2) = j, \text{ where } x_1 - x_2 = \delta \cos \theta \text{ and } y_1 - y_2 = \delta \sin \theta$$
$$\left. \text{or } x_1 - x_2 = -\delta \cos \theta \text{ and } y_1 - y_2 = -\delta \sin \theta \right\} \qquad (10)$$

The # denotes the number of elements in the set, and $M$ and $N$ are the horizontal and vertical dimensions of $I(x, y)$. Hence, by varying $\theta$ and $\delta$ multiple co-occurrence matrices can be generated for each wavelet subband. Having the co-occurrence matrix normalised to be represented as a joint probability density function, we can then in four different directions (0º, 45º, 90º and 135º) derive four second order statistical features which are most commonly used in the literature, which are: correlation, energy, dissimilarity and homogeneity, where the subband with the maximum energy is used to guide the decomposition. This best basis method would be abbreviated as BBS$_{CM}$. Moreover, the classical entropy-based best bases selection from a library of WPs proposed by Coifman and Wickerhauser [5] would be benchmarked with as well.

*C. Geometric deformation simulation*

Distortion in histopathological images tends to be different than that occurring in other imaging modalities such as in Computed Tomography (e.g. Gaussian noise) or in Ultrasound (e.g. multiplicative speckle noise). A common distortion is the white patches on slides after deparaffinisation step, which we call *craquelures* since it resembles the fine cracks in old paintings, where a technique was applied for removal of these effect prior to feature extraction as discussed in section IV-B. Another type of distortion which we like to assess its effect on the robustness of the extracted feature set is geometric deformation. Herein, an affine transformation – a shearing deformation by having non-uniform scaling in some directions – was applied to all images in the data set before feature extraction as illustrated in Fig. 8.

Each image was divided into a 4 x 4 lattice and perspective irregularities to the main descriptive features was introduced at least to two different locations in the image where the subtype prominent features (according to Table I) are mostly apparent.

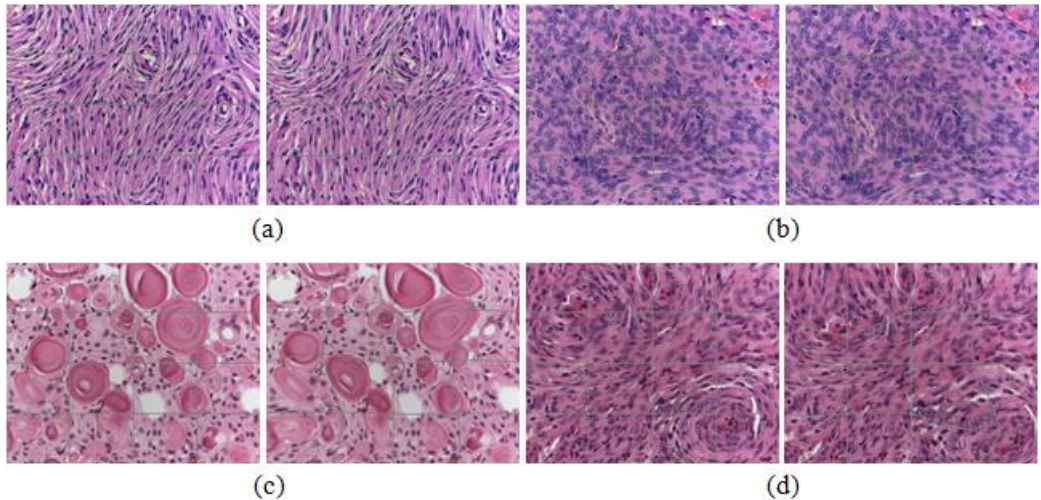

Fig. 8. An example of a meningioma subtype image subdivided into a 4 x 4 lattice and corresponding geometrically deformed image for (a) fibroblastic, (b) meningothelial, (c) psammomatos, and (d) transitional subtypes.

### D. Application to non-tissue texture images

For completeness and to demonstrate efficacy, the proposed technique (BBS$_{FD}$,) would be applied on a non-tissue dataset, where eight different and distinct texture images representing seven different classes were selected from the commonly used Brodatz texture dataset [40]. Furthermore, and based on our previous work of comparing statistical - and model based texture analysis methods [41], the BBS$_{FD}$, technique will be compared with four different texture analysis methods, which are: Gaussian Markov random field, factional Brownian motion, grey level co-occurrence matrix and run-length matrix.

## VI. OPTIMAL FEATURE CLASSIFICATION

The extracted multiresolution features from the decomposed subbands by the three aforementioned methods were trained using a classifier. Since a feature set may be performing better than another set only because its distribution is a better fit for the assumptions underlying the classifier model, the classification performance was checked using three different classifiers widely used in medical imaging to investigate the classifiers' influence on the classification accuracy.

*A. Predictive modelling*

*1) Support vector machines*: Based on the statistical learning theory and the Vapnik-Chervonenkis dimension, a support vector machine (SVM) is a kernel-based technique composed from a set of related supervised learning methods that can be used to analyse data and recognise patterns in statistical classification and regression analysis. SVM classifiers are known for their capability for controlling complexity and over-fitting, where they can generalise remarkably well in high dimensionality features vector space; however not always optimal for features with small vector spaces, and the appropriate selection of the kernel parameters might be a challenging task [42].

It is well known that the selection of the SVM kernel has a significant effect on the SVM performance. In this work a Gaussian kernel, as in (11), was chosen as it can deal with nonlinearity between class labels and corresponding attributes. Through mapping samples to a higher dimension space, it can give a similar performance to a linear kernel when sample numbers are not very large, besides numerical flexibility, it has a single parameter $\gamma$ to be calibrated [43].

$$K(x_i, x_j) = \exp\left(-\gamma \left\| x_i - x_j \right\|^2\right), \text{ for } \gamma > 0 \qquad (11)$$

A simple yet effective approach to find the best combination of the Gaussian's kernel single parameter $\gamma$ and the penalty (soft margin) parameter $C$ is by a grid-search with exponentially growing sequences of $C$ and $\gamma$, for example, $C \in \left\{ 2^{-5}, 2^{-3}, \ldots, 2^{23}, 2^{25} \right\}$; $\gamma \in \left\{ 2^{-15}, 2^{-13}, \ldots, 2^1, 2^3 \right\}$ [43]. Typically, each combination of parameter choices is checked using cross validation, and the parameters with best cross-validation accuracy are selected. For the classification of the meningioma tumours in this work, a $C$ and $\gamma$ value of $2^{23}$ and $2^{-2}$, $2^{23}$ and $2^{-12}$, $2^{17}$ and $2^{-6}$ were found to be the optimal combinations for the BBS$_{FD}$, BBS$_{CM}$, and BBS$_E$ best bases selection methods; respectively. Using these optimal parameters, the final model used for testing was trained on the whole training set. The SVM in this work was computed using the LIBSVM library [44].

*2) Naïve Bayesian classifier*: A simple – thus called naïve – Bayesian classifier (NBC) was applied also for classification. A NBC in supervised learning is considered optimal if all attributes are independent given the class. Despite the fact that this condition might not be frequent in practice, this fast and simple classifier was reported to perform well even with the presence of strong attribute dependence [45]. According to Bayesian theory and after assuming conditional independence of attributes values, NBC can be represented as:

$$P(C_i / X) = \left( \frac{P(C_i) \prod_{j=1}^{n} P(X_j / C_i)}{\sum_{i=1}^{k} P(C_i) \prod_{j=1}^{n} P(X_j / C_i)} \right) \qquad (12)$$

where $P(C_i/X)$ is the *a posteriori* probability of assigning class $i$ given feature vector $X$, $P(X/C_i)$ is the probability density function of (PDF) $X$ within the $i^{th}$ class $C_i$ for a total number of $K$ classes, $P(C_i)$ and $P(X)$ are the *a priori* probability of class $C_i$ and feature vector $X$; respectively.

*3) K-Nearest Neighbour*: The conventional *k*-nearest neighbour (kNN) classifier is a simple nonparametric classifier that operates in a given *N*-dimension feature space by finding the closest object from the training set to the one being classified based on some difference or similarity measure, such as the Euclidean distance. For a specified distance *k*, the classified object is assigned the class most frequently represented among the nearest nearby-objects, and if more than one class exist, then the classified object is assigned the class with minimum average distance to it.

The kNN meningioma classification was implemented using different values of $k = 1, 3, 5, 7,$ and 9 to search for the neighbourhood that would best characterise the cell nuclei texture among the different subtypes. Taking into consideration that the classifiers' neighbourhood should balance the trade-off between sensitivity to noise and inclusion of other information from other classes, a $k$ value of 3 was chosen empirically and after weighting the classification accuracies for each of the $k$ values, that would best capture the difference between the cell nuclei in the four-meningioma subtypes.

*B.   Dimensionality reduction*

A divergence function that can inspect for feature separability by placing no prior assumption on class-conditional densities and has a direct relation with Bayes error [50] was used for feature dimensionality reduction. The summation of the divergence measure $D_i$ for each feature $f_i$ between the four different meningioma subtypes can be defined as

$$D_i(f_i) = \sum_{k=1}^{n_c} \sum_{l>k}^{n_c} \frac{(\sigma_{k,f_i} - \sigma_{l,f_i})^2 (1 + \sigma_{k,f_i} + \sigma_{l,f_i})}{2\sigma_{k,f_i}\sigma_{l,f_i}} \qquad (13)$$

where $n_c$ is the number of subtypes – four for this work – and $\sigma_{k,f_i}$ and $\sigma_{l,f_i}$ are the standard deviation of feature $f_i$ for class $k$ and $l$; respectively.

Next the correlation between each pair of features is calculated after ranking in a descending order according to their corresponding divergence values. A threshold of ±0.8 is set for the correlation values, considering correlation values above 0.8 to be highly correlated; therefore if the absolute value of a certain correlation was found to be greater than the specified threshold, the feature with the lower divergence was excluded while the order of the remaining features is preserved.

*C. Performance estimation*

The 320 samples, which refer to 20 patients, were equally divided into four diagnostic groups (i.e. the four meningioma subtypes), each group consists of 80 samples extracted from five different patients (16 each) diagnosed with the same meningioma tumour subtype. A leave-one-patient-out approach (i.e. an external 16-folds cross-validation) was applied to validate the classification results and to avoid any possible feature selection bias [46, 47]. This is done by designing the classifier using ($n$-1) patients per subtype, where $n$ is the number of total patients each having $M$ number of sample images, and then evaluated externally on the average of the remaining set-aside single patient $M$ samples that were excluded entirely from the internal training phase. This process is repeated $n$ times covering all possible unique sets of other patients' samples; thereby an unbiased estimation is achieved.

## VII.  RESULTS

The classification performance using the SVM classifier with up to ten levels of resolution using the FD features for best basis selection (BBS$_{FD}$) is shown in Table III, where a threshold value for the FD features was not used to stop the WP decomposition. The best classification accuracy of 91.19% was achieved at the eighth level of decomposition. Alternatively, using the appropriate threshold for $D_m$ and $L_m$ as discussed in the subband optimisation section, the decomposition should terminate when there is no significant difference between the FD signatures for each subtype separately – highlighted in bold – giving an improved overall accuracy of 91.94% but with a significant save in computational time; while the best classification accuracies using the NBC and kNN were 91.01% and 76.50%. The overall classification accuracy degrades by 8.75% for SVM, 3.12% for NBC, and 9.37% for kNN if the input histopathological images were not pre-processed as in section III.B, that is, converting images to grey level instead of selecting the appropriate colour channel relative to the cell nuclei, and without applying the morphological gradient operation for reduction of mechanical distortion. Also, the geometric deformation experiments on the dataset showed a slight reduction of 0.7%, 1.5%, and 0.9% in classification performance for the SVM, NBC, and kNN classifiers, respectively.   Moreover, ensembles with bagging were applied for improving reported accuracies for the best performing classifier. Each individual SVM is trained independently using a bootstrap technique then aggregated to make a collective decision via majority voting. The accuracy of the classifier with best performance, namely SVM, and after applying the feature reduction technique in section VI-B gave a 94.12% (a 2.1% improvement in overall classification accuracy as compared to previous work [20]); also the NBC and kNN had a 1.6% and 4.85% improvement, respectively.

The selected subbands for the best bases selection process spanned by the BBS$_{FD}$ method for multiple tested images were found to be: fibroblastic (W$^1_{1,HL}$, W$^1_{2,LH}$, W$^1_{3,HH}$), meningothelial (W$^2_{1,LH}$, W$^2_{2,LH}$), psammomatous (W$^3_{1,HH}$, W$^3_{2,HH}$, W$^3_{3,HH}$, W$^3_{4,HH}$, W$^3_{5,HH}$, W$^3_{6,HH}$, W$^3_{7,HH}$, W$^3_{8,HH}$), and transitional (W$^4_{1,HH}$, W$^4_{2,HH}$, W$^4_{3,LH}$, W$^4_{4,HH}$, W$^4_{5,HH}$, W$^4_{6,HH}$, W$^4_{7,HH}$, W$^4_{8,LH}$, W$^4_{9,HH}$). It is noticed that the meningothelial subtype needed only two levels of decomposition to reach its optimum performance. The decomposition terminated at the third level for fibroblastic, at the eighth level for the psammomatous and at the ninth for transitional subtypes. A comparison is also performed to evaluate the performance of the BBS$_{FD}$ approach with two other first order statistics methods. The BBS$_{FD}$ model based method suggested in this work used the FD characteristics to guide the WP tree-structured expansion in order to construct a feature vector of the subbands having the best discriminating FD signatures. On the other hand, the statistical approaches used the highest energy for best basis selection process, where the first method (BBS$_E$) simply employed the computed highest energies of the subbands as features, and the second method (BBS$_{CM}$) extracted the co-occurrence matrix correlation, energy, dissimilarity and homogeneity (with displacement vector $\delta$ set to 1 and orientation $\theta$ in 0°, 45°, 90° and 135°) as feature for classification. The three subband decomposition approaches were also run at up to ten levels of resolution, and the corresponding classification accuracy is determined at each level. The BBS$_{FD}$ fractal approach outperformed the others, where the BBS$_{CM}$ and BBS$_E$ approaches achieved a maximum overall classification accuracy of 86.31%, 76.01% for SVM, and 83.19%, 73.50% for NBC, and 51.63%, 50.69% for kNN; respectively. Also using Wilcoxon signed-rank test, there was significance between the BBS$_{FD}$ proposed technique and the BBS$_E$ and BBS$_{CM}$ techniques ($p < 0.05$). Moreover, Coifman and Wickerhauser energy-based best bases selection from a library of WPs approach with the BBS$_{FD}$ WP tree-structured approach resulted in an increase in classification error of 5.32% for SVM, 6.26% for NBC, and 8.94% for kNN.

Table IV shows the confusion matrix for the best feature extraction method that demonstrated the highest classification accuracy (BBS$_{FD}$ classified using SVM). For comparison, the confusion matrices for the NBC and kNN are shown in Table V and VI. Also in order to assess the generalisation capacity of the SVM classifier, the number of support vectors (nSV) for each of the three different feature extraction methods is shown in Table VIII. The total nSV show that the $BBS_{FD}$ method had the least numbers for all WP decomposition levels.

Different types of texture images ranging from fine to coarse for the purpose of non-tissue texture classification were used in this paper to investigate the efficacy of the $BBS_{FD}$ technique. Eight different texture images having a size of $256 \times 256$ with 8-bit grey levels were selected from the Brodatz album. Each image that defines a separate class was divided into 32 x 32 size image segments with 50% overlapping, and a NBC was applied for classification. A holdout validation approach was used to test the accuracy, by reserving nearly one third of the image segments (64 samples) for training and the rest (192 samples) for testing. The results in Table VII show that the $BBS_{FD}$ method outperformed the texture analysis Gaussian Markov random field (GMRF) and the fractional Brownian motion (fBm) model-based methods, and the grey level co-occurrence matrix (GLCM) and run-length matrix (RLM) statistical-based methods when applied to non-tissue texture images with different types of textures.

TABLE VII
BRODATZ TEXTURE IMAGE CLASSIFICATION COMPARISON VIA FIVE DIFFERENT FEATURE EXTRACTION METHODS

| Brodatz textures | Texture feature extraction method | | | | |
|---|---|---|---|---|---|
| | GLCM | GMRF | RLM | fBm | $BBS_{FD}$ |
| D16 | 82.81% | 99.48% | 77.08% | 83.33% | 98.96 % |
| D20 | 100% | 98.96% | 100% | 80.73% | 97.40 % |
| D24 | 88.54% | 91.67% | 85.94% | 76.56% | 100% |
| D74 | 100% | 83.85% | 98.44% | 97.40% | 95.31 % |
| D93 | 99.48% | 98.44% | 88.02% | 83.85% | 99.48 % |
| D98 | 97.40% | 100% | 92.19% | 96.35% | 98.96 % |
| D106 | 97.92% | 99.48% | 98.44% | 76.04% | 100% |
| D112 | 100% | 91.67% | 89.06% | 91.67% | 100% |
| Accuracy | 95.77% | 95.44% | 91.15% | 85.74% | **98.76 %** |

TABLE III
WAVELET PACKET DECOMPOSITION USING MAXIMUM FRACTAL DIMENSION
SIGNATURE FOR BEST BASIS SELECTION VIA SVM CLASSIFIER

| Resolution | Meningioma subtype | | | | Total Accuracy |
|---|---|---|---|---|---|
| | Fibro. | Menin | Psamm. | Trans. | |
| level 1 | 87.25 | 87.00 | 92.25 | 82.75 | **87.31** |
| level 2 | 88.50 | **91.25** | 92.25 | 88.75 | **90.19** |
| level 3 | **92.25** | 90.00 | 92.25 | 86.00 | **90.13** |
| level 4 | 90.00 | 89.75 | 92.25 | 89.75 | **90.44** |
| level 5 | 91.25 | 91.25 | 91.75 | 88.75 | **90.75** |
| level 6 | 91.25 | 91.00 | 92.25 | 86.25 | **90.19** |
| level 7 | 92.50 | 91.25 | 92.25 | 86.25 | **90.56** |
| level 8 | 90.00 | 91.25 | **93.50** | 90.00 | 91.19 |
| level 9 | 86.25 | 89.75 | 93.50 | **90.75** | **90.06** |
| level 10 | 86.25 | 90.00 | 93.25 | 90.50 | **90.00** |

TABLE IV
FOUR-CLASS MENINGIOMA SVM CLASSIFICATION CONFUSION
MATRIX FOR THE $BBS_{FD}$ FEATURE EXTRACTION METHOD

| Meningioma type | | Classification | | | |
|---|---|---|---|---|---|
| | | Fibro. | Menin. | Psamm. | Trans. |
| True class | Fibroblastic | 92.50% | 03.75% | 1.25% | 02.50% |
| | Meningothelial | 03.75% | 91.25% | 00.00% | 05.00% |
| | Psammomatous | 01.50% | 01.25% | 93.50% | 03.75% |
| | Transitional | 05.50% | 01.25% | 02.50% | 90.75% |

TABLE V
FOUR-CLASS MENINGIOMA NBC CLASSIFICATION CONFUSION
MATRIX FOR THE $BBS_{FD}$ FEATURE EXTRACTION METHOD

| Meningioma type | | Classification | | | |
|---|---|---|---|---|---|
| | | Fibro. | Menin. | Psamm. | Trans. |
| True class | Fibroblastic | 83.75% | 10.00% | 00.00% | 06.25% |
| | Meningothelial | 06.25% | 88.75% | 02.50% | 02.50% |
| | Psammomatous | 01.25% | 02.50% | 92.50% | 03.75% |
| | Transitional | 10.00% | 05.00% | 03.75% | 81.25% |

TABLE VI
FOUR-CLASS MENINGIOMA kNN CLASSIFICATION CONFUSION
MATRIX FOR THE $BBS_{FD}$ FEATURE EXTRACTION METHOD

| Meningioma type | | Classification | | | |
|---|---|---|---|---|---|
| | | Fibro. | Menin. | Psamm. | Trans. |
| True class | Fibroblastic | 42.50% | 22.50% | 5.00% | 30.00% |
| | Meningothelial | 02.50% | 72.50% | 18.75% | 06.25% |
| | Psammomatous | 01.25% | 18.75% | 80.00% | 00.00% |
| | Transitional | 12.50% | 16.25% | 01.25% | 70.00% |

The corresponding kNN classifier best classifications without feature reduction for $k = 1, 3, 5, 7$, and 9, using the $BBS_{FD}$ method were 74.13%, 76.01%, 75.37%, 75.69%, and 75.69%, respectively. All were achieved in the $5^{th}$ decomposition level, except for $k = 7$ which occurred in the second level; therefore, this justifies the selection of a $k$ value of 3 for the kNN meningioma classification to ensure best performance.

Regarding classifiers behaviour illustrated in Fig. 9, the $BBS_{FD}$ classification accuracies using SVM and kNN remain fairly stable throughout all decomposition levels – improves in deeper levels of resolution – and well above compared to the other two methods. Although the $BBS_{FD}$ classification accuracies using NBC from level 2 to 4 remains higher than the other two methods, it starts to degrade afterwards, scoring accuracies equal or lower to that of the energy approach $BBS_E$.

Furthermore, the statistical measures of the classification tests in Fig. 10 and Fig. 11 show a very good sensitivity for the $BBS_{FD}$ with none of the four meningioma subtypes scoring less than 90% (the highest the psammomatous with 93.75%), the $BBS_{CM}$ had all the subtypes greater than 80% but less than 90% except for the psammomatous subtype with 92.50%, while the $BBS_E$ showed the least performance, especially for the fibroblastic subtype with 62.50%. Concerning measuring the negatives that are correctly identified, or known as specificity, again a very good performance for $BBS_{FD}$ with all subtypes above 97% (the worst was the meningothelial with 97.10%), the $BBS_{CM}$ and $BBS_E$ had a lower performance having the worst specificity of 93.75% for the transitional for the former and 90.83% for the meningothelial for the latter.

Fig. 12 shows the four meningioma subtypes classification accuracies at the best decomposition levels for all three classifiers. That is, the SVM classifier at the $8^{th}$ level, NBC at the $2^{nd}$ level, and the kNN ($k = 3$) at the $10^{th}$ level of resolution analysis. The SVM classifier outperformed the NBC and kNN counterparts, where none of the four subtypes recorded classification accuracies less than 90%. While the NBC accuracies ranged in the 80's, and the kNN was very week especially with fibroblastic subtypes.

TABLE VIII

COMPARING THE NUMBER OF SUPPORT VECTORS OBTAINED BY
BBS$_{FD}$, BBS$_{CM}$, BBS$_E$ FEATURE EXTRACTION METHODS FOR THE
FOUR-MENINGIOMA SUBTYPES AT TEN RESOLUTIONS

| Resolution | Feature extraction method | | |
|---|---|---|---|
| | BBS$_{FD}$ | BBS$_{CM}$ | BBS$_E$ |
| level 1 | 112 | 198 | 263 |
| level 2 | 71 | 126 | 251 |
| level 3 | 81 | 150 | 243 |
| level 4 | 78 | 158 | 241 |
| level 5 | 89 | 162 | 241 |
| level 6 | 81 | 179 | 245 |
| level 7 | 81 | 181 | 242 |
| level 8 | 85 | 178 | 242 |
| level 9 | 85 | 181 | 249 |
| level 10 | 82 | 179 | 243 |

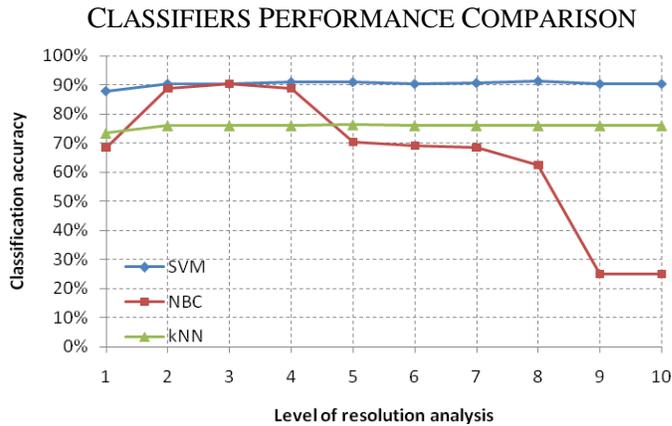

Fig. 9. Classification behaviour for SVM, NBC, and kNN classifiers using the BBS$_{FD}$ method decomposed up to the 10$^{th}$ level of resolution.

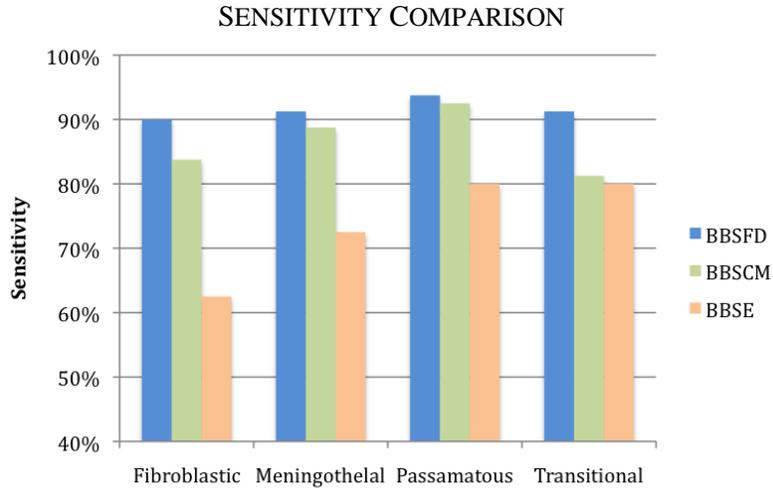

Fig. 10. Sensitivity comparison for all four meningima subtypes using the $BBS_{FD}$, $BBS_{CM}$, and $BBS_E$ methods.

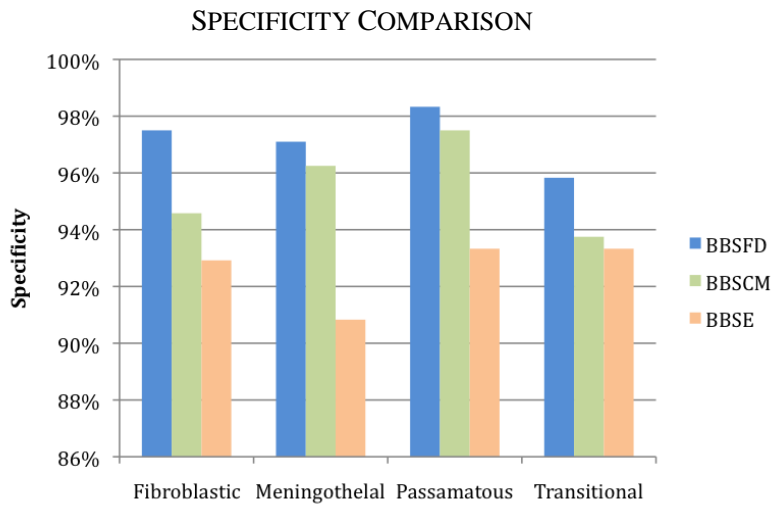

Fig. 11. Specificity comparison for all four meningima subtypes using the $BBS_{FD}$, $BBS_{CM}$, and $BBS_E$ methods.

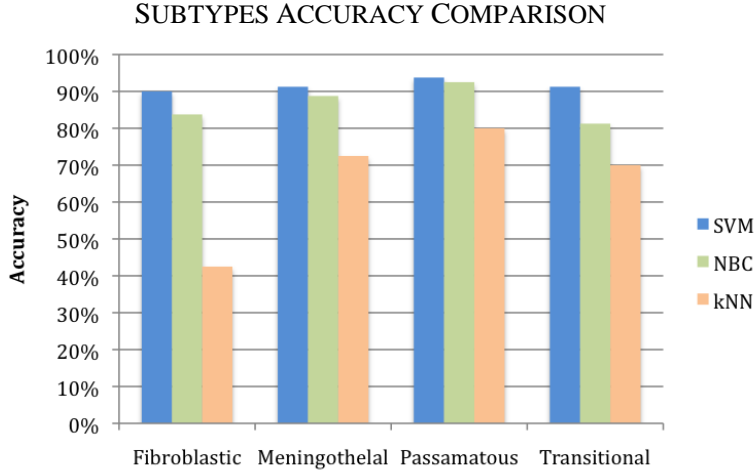

Fig. 12. Meningioma fibroblastic, meningothelial, psammomatous, and transtional subtypes accuracies using SVM, NBC, and kNN classifiers.

## VIII. DISCUSSION AND INTERPRETATION

### A. Decomposition based on subbands' fractal dimension

A different approach in assessing the fractal characteristics instead of the energy of subbands was pursued for tree-structured wavelet extension. The best basis selection process relied upon selecting the subbands with the roughest surface (i.e. highest FD) for analyses and discarding other subbands from the same level, and then the FD of the highest subbands would be used for meningioma subtypes discrimination. Results demonstrated that maximum classification accuracy was reached within two to nine resolution decompositions before starting to degrade. This was expected as the FD measures would give a reliable estimation to a certain level of resolution, whereas the more levels are decomposed the less details remains for the FD to measure; especially if the middle or high bands were selected for further analysis. Thus, determining the appropriate resolution level is not only important to save computational time but also to improve the quality of the extracted subband features. The decomposition insignificance was indicated according to the FD texture absolute difference $\mathcal{D}_m$ and lacunarity $\mathcal{L}_m$ (heterogeneity measure) between all subbands after optimising a threshold $\lambda$ that specifies how deep the image resolution can be probed. This is equivalent to excluding FD features equal to or above 2.985, considering them as nearer to noise rather than a meaningful roughness estimation of the subimage surface. Moreover, comparing the BBS$_{FD}$ technique with the statistical BBS$_{CM}$ and BBS$_E$ techniques using all three classifiers, a significant increase in improvement for the classification accuracy was achieved by 5.63% (SVM), 7.81% (NBC), 24.37% (kNN) as compared to the BBS$_{CM}$, and 18.44% (SVM), 17.50% (NBC), 25.31% (kNN) as compared to BBS$_E$. Furthermore, the reported robustness to slight geometric deformation (i.e. non-uniform scaling in some directions) could be attributed to the multiresolution capability of the proposed novel technique due to its reliance on the fractal characteristics of the examined feature texture surface - which is known for its invariance to translation, rotation and scaling [51].

### B. Computational complexity

As the decomposition levels probe further, computational cost grows higher, and the proposed BBS$_{FD}$ method works to balance this trade-off. That is, instead of decomposing several levels in search for the best classification performance and having a decomposition computational cost of $O(N\log(N))$, the BBS$_{FD}$ would terminate the search for the best bases once the wavelets subbands' absolute difference and lacunarity at a certain decomposition level is satisfied, i.e. difference between the FD become neglegible and no need to probe further. Thereby the best characterisitcs for each of the meningioma subbands would be selected for optimal classification, even though occuring at different decomposition levels (levels 3, 2, 8, and 9 for fibroblastic, meningothelial, psammomatous, and transitional subtypes, respectively), giving an optimal classificatioin accuracy of 94.12% after feature reduction.

### C. Clinical significance

On the other hand, it was expected that all three tested methods had their highest sensitivity and specificity for the psammomatous subtype, since its general morphological structure is more distinct than the others (revisit Fig. 4), and hence easier to recognise. Examining the results further, the BBS$_{FD}$ was able to outperformed the other wavelet subband decomposition methods regarding sensitivity and specificity, and in all four subtypes. More precisely, the proposed method in this work, which captures the fractal features of the underlying histopathological structure, managed to distinguish robustly between the highly similar textures of the fibroblastic and transitional subtypes that the traditional energy based best bases selection methods were weak in discrimination. This gives an indication that BBS$_{FD}$ can deal with the heterogeneity nature associated with histopathological tissue textures.

Optimal features for the meningothelial subtype were predominantly in the mid subbands, mid-high subbands for the fibroblastic, overwhelmingly in the high subbands for the psammomatous, and mainly in the high − with a few occurring in the mid − subbands for the transitional. This confirms the initial hypothesis that a multiresolution approach would be essential to deal with the heterogeneity nature and sometimes the stochastic behaviour that histopathological tissue texture could exhibit, if an optimal classification performance would be sought.

### D. Classifiers behaviour

Testing the robustness of the BBS$_{FD}$ method via three well known classifiers, results in Tables IV, V and VI containing the confusion matrices show that the SVM managed to improve the recognition of the fibroblastic meningioma subtype which is considered the most difficult among all grade I four subtypes to identify [48], as classifiers tend to miss classify it as transitional. Intuitively this improvement in the fibroblastic recognition positively paid back in the transitional subtype, where an improvement in the transitional subtype accuracy was noticed as well. The meningothelial and psammomatous subtype were relatively less complicated to characterise, showing a slightly equal performance between the SVM and NBC in classifying them; however the SVM achieved the highest accuracies in all four subtypes, and the resulting small number of support vectors support the classifiers' generalisation capacity. Best results were obtained when applying the BBS$_{FD}$ method and using the SVM classifier and after feature reduction that achieved a performance of 94.12%, while the NBC and kNN achieved an optimal accuracy of 92.90% and 79.70%, respectively.

*E. Future work*

Finally, the developed BBS$_{FD}$ technique for discriminating grade I histopathological meningiomas was based on the generalisation of the standard wavelet dyadic decomposition; however, we aim to expand the BBS$_{FD}$ to a so-called *M*-band wavelet transform, where a particular type of filter bank architecture is used to subdivide each of the dyadic or octave bands into further *M* channels. This signal decomposition into *M*, instead of 2, channels could provide more flexible tiling of the time-frequency plane, and hence achieve a better multi-scale multi-directional image filtering. Also, the accuracy of the FD estimation using least squares regression can be computed instead using the maximum likelihood method proposed in [49], and investigated if this would have an effect on the classification accuracies. Moreover, given that the opportunity to test the proposed technique on other type of histopathological tumour images was not available at this moment, the robustness of the decomposition insignificance threshold when applied to a different type of tumour needs to be investigated.

## IX. Conclusion

Dealing with the heterogeneous nature of tissue texture in a multiresolution manner while simultaneously exploiting its fractal dimension characteristics can enhance the capability of discriminating higher order statistical textural features, for which it would be otherwise difficult via ordinary human vision; and hence can reduce intra- and inter-observer variability and improve the currently followed clinical protocol in histopathological diagnosis for patient management.

A clinical decision support system that integrates a robust feature extraction method – that can deal with tissue heterogeneity – with the most appropriate classifier model design – that would improve the quality of the extracted features – for an optimised meningioma texture classification was the main concern of this work. The approach that employed the FD for wavelet tree-structured decomposition when integrated with three different classifiers demonstrated its capability to distinguish grade I histopathological meningioma images with an improved accuracy as compared to energy-based decompositions. The SVM classifier along with the BBS$_{FD}$ approach, which proved to be more statistically significant, relies on revealing texture structure complexity that would better characterise the information situated in the middle and high frequency bands. Also the appropriate decomposition levels would be detected when no more significant difference among the subbands exist, saving unnecessary computations.

## ACKNOWLEDGMENT

The author would like to thank Dr. Volkmar Hans from the Institute of Neuropathology, Bielefeld, Germany for the provision of the meningioma dataset used in this paper.